\documentclass[conference]{IEEEtran}
\IEEEoverridecommandlockouts
\usepackage[left=1.62cm, right=0.63in,top=0.75in]{geometry}

\usepackage{cite}
\usepackage{amsmath,amssymb,amsfonts}

\usepackage{graphicx}
\usepackage{textcomp}
\usepackage{xcolor}

\def\BibTeX{{\rm B\kern-.05em{\sc i\kern-.025em b}\kern-.08em
    T\kern-.1667em\lower.7ex\hbox{E}\kern-.125emX}}

\usepackage{booktabs} 
\usepackage{tabularx}

\usepackage{algorithm}
\usepackage{algpseudocode}

\begin{document}

\title{An Adaptive End-to-End IoT Security Framework Using Explainable AI and LLMs
}

\author{


\IEEEauthorblockN{Sudipto Baral}
\IEEEauthorblockA{\textit{Department of Computer Science} \\
\textit{Western University}\\
London, Canada \\
sbaral3@uwo.ca}
\and
\IEEEauthorblockN{Sajal Saha}
\IEEEauthorblockA{\textit{Department of Computer Science} \\
\textit{University of Northern British Columbia}\\
Prince George, Canada \\
sajal.saha@unbc.ca}
\and
\IEEEauthorblockN{Anwar Haque}
\IEEEauthorblockA{\textit{Department of Computer Science} \\
\textit{Western University}\\
London, Canada \\
ahaque32@uwo.ca}

}

\maketitle

\begin{abstract}
The exponential growth of the Internet of Things (IoT) has significantly increased the complexity and volume of cybersecurity threats, necessitating the development of advanced, scalable, and interpretable security frameworks. This paper presents an innovative, comprehensive framework for real-time IoT attack detection and response that leverages Machine Learning (ML), Explainable AI (XAI), and Large Language Models (LLM). By integrating XAI techniques such as SHAP (SHapley Additive exPlanations) and LIME (Local Interpretable Model-agnostic Explanations) with a model-independent architecture, we ensure our framework's adaptability across various ML algorithms. Additionally, the incorporation of LLMs enhances the interpretability and accessibility of detection decisions, providing system administrators with actionable, human-understandable explanations of detected threats. Our end-to-end framework not only facilitates a seamless transition from model development to deployment but also represents a real-world application capability that is often lacking in existing research. Based on our experiments with the CIC-IOT-2023 dataset \cite{neto2023ciciot2023}, Gemini and OPENAI LLMS demonstrate unique strengths in attack mitigation: Gemini offers precise, focused strategies, while OPENAI provides extensive, in-depth security measures. Incorporating SHAP and LIME algorithms within XAI provides comprehensive insights into attack detection, emphasizing opportunities for model improvement through detailed feature analysis, fine-tuning, and the adaptation of misclassifications to enhance accuracy.
\end{abstract}


\section{Introduction}
\label{introduction}

A rapidly evolving technological framework, the Internet of Things (IoT) is renowned for its extensive network of interconnected devices. As IoT connects the physical and digital areas, this integration has resulted in an increase in cyberattacks on IoT equipment. The challenge of protecting  IoT comes from the wide range of devices, their heterogeneity, limited processing capacity, and their pervasive accessibility. The field of IoT attack detection has evolved significantly with the emergence of machine learning (ML) and deep learning (DL) techniques \cite{9162668,sahu2021internet}. But even with these improvements, the field still encounters a number of significant challenges, which emphasizes the need for more study and innovation. A requirement for transparency in these models' decision-making processes is one of the main issues, which makes Explainable AI (XAI) necessary. The goal of XAI is to make ML/DL model outputs comprehensible to human specialists, which is essential for dependability and trust, particularly in sensitive applications where decision-making process comprehension is just as vital as decision-making itself.

Moreover, while numerous studies have proposed sophisticated models for detecting attacks in IoT networks \cite{shafiq2020corrauc, samy2020fog}, there exists a noticeable gap in the literature regarding the development of comprehensive end-to-end attack detection frameworks. Such frameworks should not only consider the development and fine-tuning of detection models but also their deployment and operationalization in real-world settings. This gap highlights the need for a holistic approach that bridges the gap between theoretical model development and practical application, ensuring that the models are not only effective in detecting attacks but also viable and user-friendly for implementation in real-world scenarios.

The goal of this research is to use XAI to improve the transparency and interpretability of ML models and provide a comprehensive, end-to-end framework for IoT threat identification. Our innovative platform facilitates a seamless transition from model development to deployment, allowing IoT devices to be effectively protected against constantly changing cyber threats. It provides a transparent security solution by combining XAI techniques like SHAP (SHapley Additive exPlanations) \cite{lundberg2017unified} and LIME (Local Interpretable Model-agnostic Explanations) \cite{ribeiro2016should} with traditional ML/DL algorithms. Its use of ML model to track network activity and detect intrusions, combined with XAI modules to clarify characteristics that are crucial for making decisions. Additionally, an intelligent Language Model Agent and Recommendation Engine provide personalized explanations facilitating mitigation response to detected threats. The main contribution of this work is as follows:

\begin{enumerate}
    \item  We integrate XAI methods with our IDS to provide clear explanations of the decision-making process, enhancing the interpretability of ML model predictions.
    
    \item  The incorporation of an Large Language Model (LLM) that adapts incident reports to the expertise level of the system administrator, ensuring effective communication and understanding.
    
    \item Proposed a complete solution from model development to deployment, ensuring that the system is not only theoretically sound but also practically applicable in real-world environments.

\end{enumerate}

This paper is organized as follows. Section \ref{sec:literature} describes the literature review of current IoT attack detection framework using machine learning models. Section \ref{sec:method} presents our proposed framework. Section \ref{sec:result} summarizes the experimentation configuration and discusses results for comparative analysis. Finally, section \ref{sec:conclusion} concludes our paper and sheds light on future research directions.

\section{Literature Review}
\label{sec:literature}
This section provides an overview of key studies focusing on the interpretability of ML/DL-based Intrusion Detection Systems (IDSs) and their constraints. Reference \cite{r1} highlights an innovative strategy by incorporating a deep neural network (DNN) for network IDS, followed by the creation of an XAI framework aimed at enhancing the DL model's clarity. The research utilizes the NSL KDD dataset to test and validate various XAI methodologies, including SHAP, contrastive explanations method, LIME, and ProtoDash. Similarly, another study mentioned in \cite{r2} adopts the SHAP method to increase the transparency of IDSs across any ML/DL-based IDS, using the NSL-KDD dataset to assess the framework's efficiency.

Reference \cite{r3} integrates XAI with an ML-based IDS to respond to adversarial attacks, starting with the construction of a random forest classifier for detecting network intrusions, then applying SHAP to elucidate and understand the outcomes of the random-forest model, with the CICIDS dataset serving for evaluation. Additionally, \cite{r4} employs a layer-wise relevance propagation (LRP) technique to assess the importance of input features and provide feedback to users, aiding them in identifying the features that significantly influence the IDS's predictions. In \cite{r5}, a novel explanation method is introduced to address misclassifications by ML/DL-based IDSs, identifying necessary adjustments for accurate data sample classification and identifying critical features that explain the misclassification, verified using the NSL-KDD dataset.

A major shortcoming of the existing research literature is the absence of comprehensive end-to-end frameworks that combine XAI with ML/DL models to ensure that the decisions made by these systems are transparent, interpretable, and actionable, while also utilizing their strengths for effective attack detection. Moreover, despite the fact that numerous XAI approaches have been studied to provide insight into model decision-making processes, comparatively little research has been done to operationalize these technologies in actual IoT environments. This gap emphasizes how important it is to have frameworks that connect the theoretical and practical sides of cybersecurity solutions in order to guarantee that they are not only efficient at identifying threats but also easily adaptable and user-friendly for a wide range of IoT applications. 

\begin{figure*}[!htbp]
    \centering
    \includegraphics[width=\textwidth]{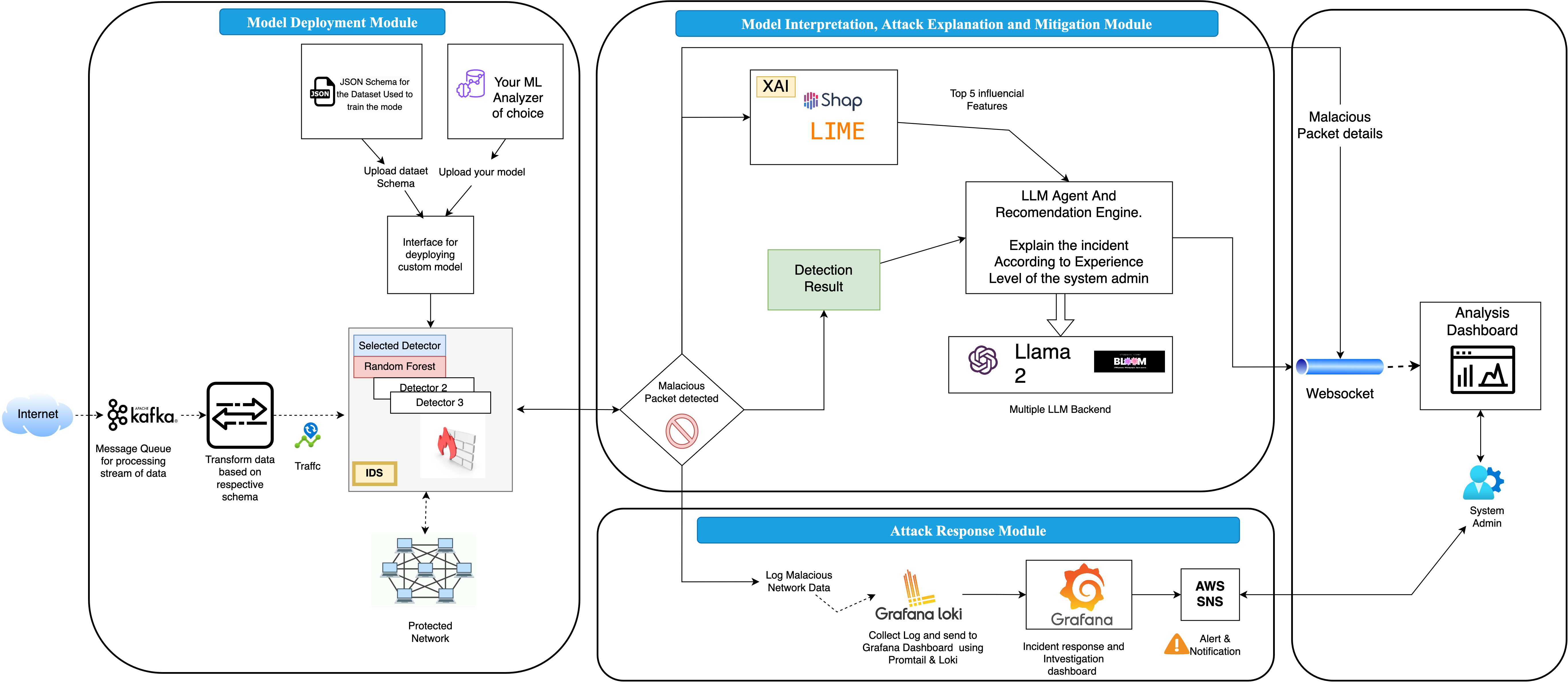}
    \caption{High-level Overview of the Proposed NIDR Framework using XAI and LLMs}
    \label{fig:methodology}
\end{figure*}

Motivated by shortcomings we found in the literature, we propose a new, end-to-end framework for real-time attack detection and mitigation. This framework distinguishes itself by combining XAI with IDS resulting in understandable and transparent explanations of the decision-making processes. This integration addresses a critical shortcoming in previous research by significantly improving the interpretability of ML/DL model predictions. Furthermore, our approach presents an LLM Agent, a novel feature that customizes incident reports based on the system administrator's skill level. Our framework offers a comprehensive solution from model generation to deployment, with a strong emphasis on practical applicability in real-world IoT scenarios. Furthermore, we integrate logging and visualization capabilities into our platform to provide an extensive and intuitive attack response and investigation dashboard, improving operational effectiveness. This work aims to substantially advance the field of IoT security by addressing these important gaps and using our creative contributions, offering reliable, understandable, and visible ways to defend against cyberattacks in IoT environment. Our research focuses on closing the gaps identified in the current literature and providing new guidelines for similar studies in the future.

\section{Proposed Framework}
\label{sec:method}


Our proposed Network Intrusion Detection and Response Framework (NIDRF) depicted in Fig. \ref{fig:methodology} offers a holistic, adaptive, and transparent security solution. Unlike existing works, our framework integrates real-time monitoring, advanced intrusion detection leveraging XAI, and an automated response mechanism into a unified system. Our proposed framework consists of four main modules such as: \textit{Model Deployment Module, Model Interpretation Module, Attack Explanation and Mitigation Module, and Attack Response Module}. In this section, we discuss each component in detail.


\begin{algorithm}
\caption{Custom ML Model Deployment}
\label{algo:model}
\begin{algorithmic}[1]
\Procedure{ModelInferenceWorkflow}{}
    \State model $\gets$ DeployModel(ReceiveModelUpload())
    \State schema $\gets$ DefineSchema(ReceiveSchemaDefinition())
    \State kafkaStream $\gets$ InitializeKafka("InputData", "OutputPredictions")
    \While{\textbf{true}}
        \State data $\gets$ kafkaStream.Consume()
        \If{Validate(data, schema)}
            \State kafkaStream.Publish(model.Predict(
            \State \hspace{\algorithmicindent}Preprocess(data, schema)))
        \EndIf
    \EndWhile
\EndProcedure
\end{algorithmic}
\end{algorithm}


\subsection{Model Deployment Module}
Our proposed framework features a flexible and adaptable architecture designed to seamlessly integrate and deploy various ML models. Deployment process is summarized in Algorithm (\ref{algo:model}). This architecture is model-agnostic, meaning it can work with any custom-built detection model provided through an Application Programming Interface (API). This approach guarantees that the framework can easily accommodate different detection models for analysis without requiring significant changes to the framework itself. The model deployment process unfolds in two distinct stages as below:

\begin{enumerate}
    \item \textbf{Data Transformation and Processing:} Initially, model weights are uploaded using the API, followed by the submission of the JSON schema corresponding to the dataset utilized for model training. Based on the provided JSON schema, a transformer is instantiated. This transformer facilitates the conversion of raw internet traffic data into a format intelligible to the ML model in use.
    
    \item \textbf{Integration of Message Queue Service:} Handling raw traffic streams within the architecture involves the integration of a message queue service, a robust distributed system serving as the foundation for managing data flows. Incoming internet traffic is aggregated into the message queue upon entering the network. Subsequently, the IDS system functions as a consumer of this queue, extracting data for detection purposes, with the transformer operating intermediary in this process.
\end{enumerate}

\subsection{Model Interpretation Module}
This module is a key component of our proposed framework, designed to enhance transparency and build trust by clarifying machine learning predictions. Utilizing SHAP and LIME algorithms, it explains the rationale behind predictions, providing network administrators with actionable insights for informed decision-making. Our framework provided a dual-layered interpretation as below:

\begin{enumerate}
    \item \textbf{Global and Feature-level Explanations with SHAP:} Firstly, we utilized SHAP, leveraging its game-theoretic approach to explain the output of any machine learning model by quantifying the contributions of each feature using Shapley values.
    
    \item \textbf{Local Insights with LIME:} Complementing SHAP, we used LIME to provide a local understanding of predictions, focusing on local fidelity to accurately reflect the classifier's behavior ``around" the instance being predicted. 
\end{enumerate}

In addition to the explanation, this module also provides a mechanism to identify the top five influential features by analyzing the SHAP and LIME values. XAI algorithms, such as SHAP  and LIME return a numerical value representing the influence of individual features on the prediction made by a machine learning model. A positive value for a feature indicates that its presence or increase contributes positively to the model's prediction, while negative values suggest a negative impact. To determine the top 5 influential features, we selected those with the highest absolute values. Initially, we generated two sets of features, each comprising the top five influential features recommended by SHAP and LIME, respectively. Typically, some features appear in both sets. Hence, as depicted in Equation \ref{eq:final_features}, the final list of features is obtained by combining the top features recommended by SHAP and LIME.

\begin{equation}
F_{\text{final}} = F_{\text{SHAP}} \cup F_{\text{LIME}}
\label{eq:final_features}
\end{equation}

\subsection{Attack Explanation and Mitigation Module}
Our framework enhances the comprehensibility of detected network attacks by translating complex machine learning outputs into easily understandable narratives and possible mitigation techniques. This process, powered by LLM, is structured into three main sub-modules as below:

\begin{enumerate}
    \item \textbf{Prompt Generation:} Our prompt generator accommodates four input values: Influential Features, Predicted Attack, Experience Level, and Descriptiveness. Influential Features and Predicted Attacks are internally processed by the prompt generator, while users can manipulate descriptiveness and experience level through the dashboard interface. Consequently, a prompt is dynamically generated based on the received values.
    
    \item \textbf{LLM Integration:} A seamless integration with the LLM backend is pivotal for transforming these prompts into detailed explanations. We establish a secure and efficient communication pathway with the LLM provider through a configuration file. These configurations, which include essential parameters such as the API URL and authentication keys, are integrated into the framework via environment variables.
    
    \item \textbf{Response Reception and Dissemination:} After providing prompts to the connected LLM provider, the generated response is obtained through HTTP communication. This response is then relayed to the system administrator via the Analysis Dashboard, which receives real-time updates through WebSocket communication. Such explanations empower administrators to swiftly grasp the nature of detected threats, facilitating informed decision-making for subsequent actions.
    
\end{enumerate}

\subsection{Attack Response Module}
Our framework integrates a comprehensive attack response module designed to streamline the detection, analysis, and response to network security incidents. This module integrates three critical functionalities: logging, monitoring, and alerting, providing a robust system for network security management.

\begin{enumerate}
    \item \textbf{Logging with Grafana:} We employ Grafana, a leading open-source platform for analytics and monitoring, to handle our logging needs. Grafana enables the detailed logging of network activities, including packet values and timestamps, facilitating in-depth analysis of network traffic and potential security breaches. 
    
    \item \textbf{Alerting with AWS SNS:} For alerting, our system integrates with Amazon Simple Notification Service (AWS SNS), a flexible, fully managed messaging service. This integration allows for the efficient dissemination of alerts and notifications in response to detected threats.

    \item \textbf{Real-Time Monitoring and Control:} The core of our incident response module is directly linked to our framework's API via WebSocket, ensuring real-time monitoring and centralized control over the IDS. This setup enables immediate notification and visualization of detected attacks through our user-friendly dashboard, enhancing situational awareness and facilitating rapid response actions.
\end{enumerate}





\begin{table*}[!htbp]
\centering
\caption{Summary of Top 5 SHAP and LIME Features for Ten Test Sample (TS)}
\label{table:shapvslime}
\begin{tabularx}{\textwidth}{|c|X|X|c|c|}
\hline
\textbf{TS Index} & \textbf{Top 5 SHAP Features} & \textbf{Top 5 LIME Features} & \textbf{Actual Class} & \textbf{Predicted Class} \\
\hline
TS 1 & Header\_Length, DNS, syn\_flag\_number, Rate, ece\_flag\_number & Protocol, Type, Number, ICMP, Weight, fin\_count & DoS-TCP\_Flood & DoS-TCP\_Flood \\
\hline
TS 2 & Header\_Length, psh\_flag\_number, DNS, Srate, Rate & DNS, Weight, Protocol, Type, fin\_flag\_number, Number & DDoS-SYN\_Flood & DDoS-SYN\_Flood \\
\hline
TS 3 & Header\_Length, Rate, fin\_flag\_number, syn\_count, Drate & DNS, Protocol, Type, Number, syn\_flag\_number, fin\_flag\_number & DDoS-ICMP\_Fragmentation & DDoS-ICMP\_Fragmentation \\
\hline
TS 4 & Header\_Length, Drate, syn\_flag\_number, ack\_count, Rate & DNS, Protocol, Type, Weight, syn\_flag\_number, psh\_flag\_number & DDoS-ICMP\_Flood & DDoS-ICMP\_Flood \\
\hline
TS 5 & Header\_Length, rst\_flag\_number, syn\_flag\_number, Rate, HTTPS & Protocol, Type, DNS, Number, Weight, syn\_flag\_number & DDoS-RSTFINFlood & DDoS-RSTFINFlood \\
\hline
TS 6 & Header\_Length, DNS, Drate, HTTP, syn\_flag\_number & DNS, psh\_flag\_number, syn\_flag\_number, Weight, fin\_flag\_number & DoS-SYN\_Flood & DDoS-ICMP\_Fragmentation \\
\hline
TS 7 & Header\_Length, Drate, DNS, syn\_flag\_number, Rate & ICMP, Protocol, Type, rst\_flag\_number, Number, Magnitue & DoS-UDP\_Flood & Mirai-greeth\_flood \\
\hline
TS 8 & Header\_Length, Drate, DNS, ack\_count, syn\_flag\_number & Weight, rst\_flag\_number, syn\_flag\_number, Protocol, Type, AVG & DDoS-UDP\_Flood & Mirai-greeth\_flood \\
\hline
TS 9 & Header\_Length, Drate, DNS, syn\_flag\_number, Rate & Protocol, Type, Number, psh\_flag\_number, syn\_flag\_number, Weight & DoS-UDP\_Flood & Mirai-greeth\_flood \\
\hline
TS 10 & Header\_Length, Drate, syn\_flag\_number, Rate, HTTPS & Weight, psh\_flag\_number, rst\_flag\_number, fin\_count, fin\_flag\_number & DDoS-ICMP\_Flood & DDoS-ICMP\_Fragmentation \\
\hline
\end{tabularx}
\label{tab:shap-lime-summary}
\end{table*}

\section{Experiment and Result Analysis}
\label{sec:result}

\subsection{Dataset preperation}
In our experimental investigation, we employed the CIC-IOT-2023 \cite{neto2023ciciot2023} attack dataset, a recent repository featuring a spectrum of 33 distinct IoT attack scenarios. To construct our experimental dataset, we integrated nine instances of attack data with benign traffic data. Specifically, we selected the following attack types as our focal points: DDoS-ICMP Flood, DDoS-UDP Flood, DDoS-TCP Flood, DDoS-PSHACK Flood, DDoS-SYN Flood, DDoS-RSTFINFlood, DDoS-SynonymousIP Flood, DoS-UDP Flood, and DoS-TCP Flood. We ensured a minimum of 20,000 data points in each category.

\subsection{Model Interpretability Analysis using SHAP and LIME}
To assess the performance of the interpretation module of our proposed framework, we trained a random forest classifier using the CIC-IOT-2023 \cite{neto2023ciciot2023} attack dataset. The resulting model demonstrated strong performance, achieving approximately 99.97\% accuracy during testing. For further analysis of the model performance using XAI and LLM, we uploaded our trained model into the framework through the model deployment module. 

According to Table \ref{table:shapvslime}, both SHAP and LIME highlight network traffic characteristics such as ``Header\_Length,'' ``DNS,'' and various flag numbers (e.g., ``syn\_flag\_number,'' ``fin\_flag\_number'') as crucial for classifying network attacks. This highlights a foundational principle in network security: the anomalies in network traffic patterns are indicative of malicious activities. For example, unusually long headers or a high rate of specific flag usage can signal an ongoing attack. The variation in feature importance between SHAP and LIME methods reflects their underlying mechanisms---global versus local interpretability. This discrepancy is particularly helpful, as it suggests that while some features consistently influence the model's predictions across different instances (SHAP), others may only be significant in specific contexts (LIME). This duality emphasizes the complexity of network attacks and the need for adaptive, context-aware models in IDS.

Analyzing misclassified entries, we notice that certain features appear uniquely in either SHAP or LIME lists, which may point to why the model struggles with certain predictions. For example, ``Drate'' and ``HTTP'' are identified by SHAP for test observation with index 6, but LIME emphasizes ``psh\_flag\_number'' and ``fin\_flag\_number,'' suggesting a possible misalignment in how the model weights protocol-specific features versus flag behaviors in its decision-making process. This observation implies that for misclassified instances, the model might be relying on features that, while generally important, fail to capture the nuanced differences required to distinguish between similar attack types, such as differentiating between DoS and DDoS attacks or identifying specific attack strategies like Mirai-greeth\_flood.

In cases where SHAP and LIME agree on the importance of certain features (e.g., ``Header\_Length,'' ``DNS''), yet misclassification occurs, it may indicate that these features, despite their overall importance, are insufficient by themselves to accurately classify all instances. This suggests the need for either additional features or a re-evaluation of how these features are processed and weighted by the model. The analysis of misclassifications through the lens of SHAP and LIME can guide model refinement. For example, extending the model with additional context-specific features or adjusting the weighting of features based on insights from both SHAP and LIME could address the misalignments leading to misclassifications. Additionally, the comparative analysis highlights the importance of incorporating both global and local interpretability perspectives for a more holistic understanding of model behavior. This dual perspective can reveal discrepancies in feature importance that, if addressed, could significantly improve model accuracy, especially in complex scenarios where subtle distinctions between classes are crucial.

\begin{table*}[!htbp]
    \centering
     \caption{Comparison of different LLM providers on a sample test data}

    \begin{tabular}{|c|p{2cm}|p{2cm}|p{4cm}|p{4cm}|}
    \hline
    \textbf{Prediction} & \textbf{Features} & \textbf{Prompt} & \textbf{Gemini Response} & \textbf{OPENAI Response} \\ \hline
    DoS-TCP\_Flood & Header\_Length, DNS, syn\_flag\_number, Rate, ece\_flag\_number, psh\_flag\_number, Protocol, syn\_flag\_number, ICMP & A DoS-TCP\_Flood Attack is detected by our Intrusion detection system. The top influential features for detecting the attack according to SHAP and LIME are given below. Explain the influential features and give a brief mitigation plan. Keep it concise & \textbf{Explanation:} DoS-TCP Flood Detected (SHAP/LIME) Influential Features: eader\_Length: Abnormally large headers consuming resources. syn\_flag\_number: A high number of SYN flags (TCP connection initiation) points to unfinished connections. DNS \& ICMP: These protocols typically use UDP, not TCP. Traffic on these ports suggests spoofed packets. Flags (ece, psh): Rarely used flags in legitimate TCP traffic.
    
    \textbf{Mitigation:} Rate Limit SYN Packets: Restrict incoming TCP connection attempts. Filter Spoofed Packets: Block traffic on unexpected protocols (DNS/ICMP) on TCP ports. & \textbf{Explanation:} Header Length: Indicates abnormal packet structures. DNS: Potential target for DDoS attacks. SYN Flag Number: SYN flood attack detection. Rate: A high data transfer rate may indicate a flood attack. ECE Flag Number: Potential exploit attempts. PSH Flag Number: Push flood attack indicator. Protocol: TCP is commonly targeted in DoS attacks. ICMP: This may indicate ICMP flood attacks.
    
    \textbf{Mitigation:}Implement SYN request rate-limiting. Configure firewalls to drop abnormal packets. Use intrusion prevention systems to detect suspicious traffic. Employ traffic analysis tools to monitor and filter abnormal rates. Keep network devices updated and deploy DDoS mitigation services.
    \\
    \hline
    \end{tabular}%
    
    \label{tab:llm-explanations}
\end{table*}

\subsection{Model Interpretability Analysis using Gemini and OPENAI}
Both LLMs explained critical features associated with the DoS-TCP\_Flood attack, such as `Header\_Length', `DNS', `syn\_flag\_number', `Rate', and TCP-specific flags (`ece\_flag\_number', `psh\_flag\_number'). These features are indicative of abnormal network behaviors and patterns, such as oversized packet headers or an excessive number of SYN flags, which are typical markers of DoS attacks. Each model proposed mitigation strategies aligned with standard cybersecurity practices, including rate-limiting SYN packets, filtering spoofed packets, implementing SYN cookies, and utilizing traffic shaping techniques. These strategies reflect a deep understanding of network security mechanisms and the specific nature of DoS attacks.

In a comparative analysis of responses from Gemini and OPENAI to a cybersecurity prompt, each LLM showcased distinctive strengths in addressing the detection and mitigation of a DoS-TCP\_Flood attack. According to Table 2, Gemini's response was notably concise, offering a clear overview of the key features indicative of the attack, as identified by SHAP and LIME methodologies. It highlighted the necessity for targeted countermeasures such as filtering spoofed packets and implementing rate limiting, directly aligning mitigation strategies with the attack's specific characteristics. In contrast, OPENAI provided a more detailed exposition, both in terms of feature explanation and the range of suggested mitigation strategies. The approach advocated by OPENAI emphasized a comprehensive, multi-layered security framework, incorporating a variety of defenses from firewall configurations to intrusion prevention systems and sophisticated traffic analysis tools. This contrast between the two models highlights a pivotal distinction in their application: while Gemini focuses on succinct, direct responses tailored to the immediate features of an attack, OPENAI opts for a broader, more detailed perspective, advocating an extensive suite of security measures to combat potential threats.

\section{Conclusion}
\label{sec:conclusion}
This paper presented a novel end-to-end framework for IoT security that integrates Explainable AI (XAI) and Large Language Models (LLMs) for enhanced attack detection and response. By integrating XAI techniques like SHAP and LIME with an intelligent LLM Agent, the framework ensures the interpretability of ML model predictions. Empirical tests using the CIC-IOT-2023 dataset validated the effectiveness of our proposed framework in detecting and mitigating threats, proving its adaptability and potential in safeguarding IoT network. Future research will focus on expanding the framework’s detection capabilities and exploring emerging XAI methods for deeper insights, aiming to strengthen IoT security against evolving threats.

\end{document}